\documentclass[utf8]{FrontiersinHarvard} 

\usepackage{url,hyperref,microtype,subcaption}
\usepackage[onehalfspacing]{setspace}
\usepackage{mathtools,leftidx}

\def\keyFont{\fontsize{8}{11}\helveticabold }
\def\firstAuthorLast{Gerdpratoom {et~al.}} 
\def\Authors{Nuthasith Gerdpratoom and Kaoru Yamamoto}
\def\Address{Graduate School and Faculty of Information Science and Electrical Engineering, Kyushu University, Fukuoka, Japan}

\begin{document}
\onecolumn
\firstpage{1}

\title[NMPC-Based Flocking with Local Obstacle Avoidance]{Decentralized Nonlinear Model Predictive Control-Based Flock Navigation with Real-Time Obstacle Avoidance in Unknown Obstructed Environments  } 

\author[\firstAuthorLast ]{\Authors} 
\address{} 

\extraAuth{}

\maketitle
\begin{abstract}

\noindent
This work extends our prior work on the distributed nonlinear model predictive control (NMPC) for navigating a robot fleet following a certain flocking behavior in unknown obstructed environments with a more realistic local obstacle avoidance strategy. More specifically, we integrate the local obstacle avoidance constraint using point clouds into the NMPC framework. Here, each agent relies on data from its local sensor to perceive and respond to nearby obstacles. A point cloud processing technique is presented for both two-dimensional and three-dimensional point clouds to minimize the computational burden during the optimization. The process consists of directional filtering and down-sampling that significantly reduce the number of data points. The algorithm's performance is validated through realistic 3D simulations in Gazebo, and its practical feasibility is further explored via hardware-in-the-loop (HIL) simulations on embedded platforms.

\tiny
 \keyFont{ \section{Keywords:} Nonlinear MPC, flocking, local obstacle avoidance, hardware-in-the-loop, distributed control} 
\end{abstract}

\section{Introduction}

In the past few decades, multi-agent systems have gained much attention in science and engineering due to the advancement of digital computers and the advantages of mimicking group behaviors to accomplish complex tasks. Our study focuses on implementing flocking behavior for a robot fleet inspired by groups of birds and fish in nature. \cite{reynold} introduced the boid model to simulate bird flocking with cohesion, separation, and alignment. Many researchers have utilized Reynolds' boid model from a control perspective, for example, \cite{reynold_paper1}, \cite{reynold_paper2}, \cite{reynold_paper3}, \cite{reynold_paper4}, and \cite{reynold_paper5}. Recently, more advanced strategies have been proposed, employing optimization-based control, such as model predictive control (MPC), for flocking. This control scheme involves agents predicting future states and adjusting control actions based on minimizing cost function, offering benefits in handling complex problems, addressing state constraints, and enabling smoother control actions over longer prediction horizons (see \cite{mpc}). Nonlinear model predictive control (NMPC) provides more flexibility than linear MPC, accommodating complex system dynamics and constraints without compromising nonlinearities.

Recent developments in optimization-based control have addressed challenges in multi-robot systems. One of them is the work done by \cite{drone_mpc}. The authors proposed MPC problems for the leader-follower structure of unmanned aerial vehicles (UAVs). The leader UAV's problem was cast to track the specified trajectory, and a fully nonlinear UAV model was utilized to constrain the optimization. Meanwhile, the followers' problem was retaining the formation among the group of UAVs with limited neighbor information to ease inter-UAV communication. The model enforced for the followers was a simplified two-layer uncoupled model, including translational and rotational motion, to reduce the computational burden. The authors implemented MPC for the followers in the linear parameter varying (LPV) fashion, LPV-MPC. \cite{fixed-wing_flocking} proposed an NMPC strategy specifically for fixed-wing UAVs' flocking. The authors presented a three-dimensional flocking model based on the distributed NMPC, encoding flocking rules as terms in the objective function. The NMPC was then solved using a nonlinear programming solver in the CasADi optimization framework (see \cite{casadi}). Safe navigation has become a hot topic in robotics and control in recent years. \cite{cbf_1} have proposed a distributed controller for multi-robot's safe navigation. The authors characterized obstacle and collision avoidance by leveraging control barrier functions (CBFs) condition as optimization constraints. While the presented control scheme did not explicitly consider predicted future states and inputs like in predictive controls, the forthcoming system's behavior was captured concerning the specified safe set. \cite{dnmpc_cbf} integrated exponential CBFs (ECBFs) into a decentralized NMPC scheme for safely controlling multi-quadcopters under thrust constraints and limited detection range situations. This control strategy enhances the reliability of CBF-based safety conditions by incorporating them within a receding horizon framework. The authors derive both conservative and practical bounds on the detection range required to preserve ECBF-based safety guarantees. The optimization-based control strategies for navigating groups of robots in obstructed environments appear effective. However, most studies assumed that obstacles were well-defined. This paper aims to close the gap between these strategies and real-world robotics applications.

This work builds upon prior studies on NMPC-based flock navigation, particularly those presented by \cite{aneek1} and \cite{aneek2}. The authors have introduced the NMPC-based flock navigation strategy with modified flocking rules consisting of graph-distance hierarchy and cohesion/alignment dynamic trade-off to navigate the fleet smoothly through an obstructed environment. The NMPC formulation allows one to explicitly impose each rule in the optimal control formulation, and it was successfully evaluated via numerical simulation and laboratory experimentation. However, in these works, it was assumed that an analytic expression of the obstacles (as a set of smooth inequalities proposed by \cite{PANOC2}) becomes available to each agent as soon as they are in their detection range. In other words, the obstacles were well-defined and did not reflect the obstacle representation in the actual robotics implementation. Therefore, this work further develops the theoretical foundations by investigating such practical aspects, which were previously neglected. In particular, we consider the situation in which the obstacles are perceived by local onboard sensors such as LiDAR and incorporate the whole process of obstacle detection, data suppression, and obstacle avoidance. We focus specifically on point cloud data, which represents the common information type for mobile robots' perception of the environment. The obstacle avoidance algorithm based on point cloud has been well studied. The most widely used algorithm is the artificial potential field proposed by \cite{potential1}, which has been utilized in many mobile robot projects, such as the collision-free backup controller for a micro aerial vehicle based on 2D point cloud, see \cite{potential3}, or decentralized control of a robot group for manipulation tasks, see \cite{potential2}. Another reactive approach is vector field histogram (VFH), which maps the radial point cloud information in a polar coordinate into a histogram and selects the obstacle-free direction for the robot to steer to, see \cite{vfh}. In this study, we incorporate obstacle information in point cloud data type into the optimal control formulation, adopting a strategy to minimize the amount of data input required by the solver to reduce the computational burden for both two and three-dimensional point clouds. The problem is solved using the Proximal Averaged Newton-type method for Optimal Control (PANOC), as introduced by the authors of \cite{PANOC} via OpEn (Optimization Engine) code generation in Rust, developed in \cite{OpEn}. The study is evaluated through a close-to-real-world simulation, Gazebo (see \cite{gazebo}), where physical uncertainties are considered. The algorithm is executed in a distributed manner asynchronously with the ROS (Robot Operating System) framework, and the robot fleet navigates safely in an obstructed environment.

Differently from the experiment setup of \cite{aneek2}, in which the NMPC algorithm was executed in a decentralized manner on a station computer, providing optimal control actions to each robot, we further study the feasibility of implementing the algorithm in the embedded platforms, where computational resources are limited, through hardware-in-the-loop (HIL) simulation, see \cite{hil1}, \cite{hil2}. We connect Raspberry Pis, widely used in low-cost robotics-embedded platforms, to a computer running a realistic simulation. The NMPC algorithm is run in a more distributed fashion inside the target hardware while simultaneously monitoring computational loads and assessing trajectory quality.

\subsection{Contributions}
In this work, there are two main contributions. First, we introduce obstacle avoidance based on local environmental information that fits our original NMPC formulation. We also propose point cloud processing that can significantly reduce the computational burden for both two-dimensional and three-dimensional point clouds. Second, we conduct an HIL simulation to investigate the feasibility of implementing NMPC-based flock navigation on Raspberry Pi 4 in a fully distributed manner.
\subsection{Outline}
Section 2 briefly outlines the NMPC-based flock navigation with modified flocking rules, serving as a self-contained context for readers. Section 3 discusses the main results of this work, consisting of point cloud data processing, obstacle avoidance constraints, and NMPC formulation featuring a local obstacle avoidance strategy. Section 4 presents the evaluation of the algorithm through realistic simulation scenarios, as well as an HIL simulation conducted on embedded platforms.

\section{Preliminaries}

In this section, we will briefly explain our prior work on distributed NMPC-based flock navigation with modified flocking rules. The optimal control problem formulation and modified flocking rules will be discussed in the following subsections. A detailed explanation can be found in the works done by \cite{aneek1} and \cite{aneek2}.

\subsection{Setting}

A system consisting of $N$ agents in an $n_{p}$-dimensional space, categorized into leaders and followers, is considered. Leaders are given trajectories, while followers only react to immediate surroundings and lack knowledge of any predefined destination. At time step $t$, let ${\mathcal{N}}_{i}^{t}$ be the index set of neighbors, including itself (agent $i$) and $ \overline{\mathcal{N}}_{i}^{t}$ be the index set of neighbors without itself. Let $x_{i}^{t} \in {\mathbb{R}}^{n}$ be the state vector of agent $i$, and $y_{i}^{t}$ be the vector that contains agent $i$'s position $p_{i}^{t}\in {\mathbb{R}}^{n_{p}}$ and velocity $v_{i}^{t}\in {\mathbb{R}}^{n_{p}}$ in the global frame defined as $y_{i}^{t}=[p_{i}^{t^{\top}},v_{i}^{t^{\top}}]^{\top}$. The variables sequence defined along the prediction horizon $T$ are bold-faced. For instance, the sequence of the control input $u_{i}^{t} \in {\mathbb{R}}^{n_{u}}$, computed at time $t$, is $\textbf{\textit{u}}_{i}^{t}:=u_{i}^{t|t} ... \thinspace u_{i}^{t+T-1 |t}$. Similarly, the sequence of the predicted state of agent $i$ is $\textbf{\textit{x}}_{i}^{t}$. The nonlinear discrete-time state equation for each agent $f_{i}: {\mathbb{R}}^{n} \times {\mathbb{R}}^{n_{u}} \rightarrow {\mathbb{R}}^{n}$ can be described as follows
\begin{equation}
    x_{i}^{t+k+1|t} =f_{i} (x_{i}^{t+k|t} ,u_{i}^{t+k|t} ),\quad k=0,...,T -1.
\label{eq:state}
\end{equation}
The goal is for followers to find optimal actions $u_{i}^{t |t}$ for navigating through obstructed environments while minimizing deviation from the proposed flocking rules, thus ensuring fleet connectivity.

\subsection{Objective Function}

To achieve our main objective, we introduced a quadratic minimum-effort cost function given by
\begin{equation}
    J(\textit{\textbf{u}}_{i}^{t}) =\| \textit{\textbf{u}}_{i}^{t} \| _{R_{i}}^{2} +\sum _{k=0}^{T -1} \gamma ^{k} \| y_{i}^{t+k+1|t} -\overline{y}_{i}^{t+k+1|t} \| _{Q_{i}^{t}}^{2},
\label{eq:cost}
\end{equation}
where $R_{i} \in {\mathbb{R}}^{n_{u} \times n_{u}}$, and $Q^{t}_{i} \in {\mathbb{R}}^{2n_{p} \times 2n_{p}}$ are positive semi-definite diagonal weight matrices corresponding to the first and second term. Let $y_{j|i}^{t}$ be the output of agent $j \in  \overline{\mathcal{N}}_{i}^{t}$, detected by agent $i$ at time instant $t$, and  $\overline{y}_{i}^{t+k|t}=[\overline{p}_{i}^{t+k|t^{\top}},\overline{v}_{i}^{t+k|t^{\top}}]^{\top}$ be the weighted average of $y_{j|i}^{t+k|t}$, where
\begin{equation}
\begin{aligned}
\quad \overline{p}_{i}^{t+k|t} &= \sum_{j\in {\mathcal{N}}_{i}^{t+k|t}} w_{{\rm p},j|i}^{t+k|t} p_{j|i}^{t+k|t}, \\
\quad \overline{v}_{i}^{t+k|t} &= \sum_{j\in {\mathcal{N}}_{i}^{t+k|t}} w_{{\rm v},j|i}^{t+k|t} v_{j|i}^{t+k|t}, 
\end{aligned}
\label{eq:y_bar}
\end{equation}
for $k=0,\ldots,T -1$. The weights $w_{{\rm p},j|i}^{t+k|t}$ and $w_{{\rm v},j|i}^{t+k|t}$ will be discussed in the modified flocking rules section (Section \ref{modified_flocking}). The errors of the predicted states can be accumulated during the prediction. Hence, a discount factor ${\gamma}\in (0,1]$ is introduced by prioritizing the near-future prediction.

\subsection{Optimization Constraints}
\subsubsection{State and Input Constraints}
The input and state constraints, excluding separation and obstacle avoidance, are assigned as
\begin{equation}
\quad x_{i}^{t+k+1|t} \in {\mathcal{X}}_{i}\quad\text{and}\quad u_{i}^{t+k|t} \in {\mathcal{U}}_{i},
\label{eq:state_input_constraint}
\end{equation}
where ${\mathcal{X}}_{i} \subset {\mathbb{R}}^{n}$ and ${\mathcal{U}}_{i} \ \subset {\mathbb{R}}^{n_u}$ are the feasible sets for the states and inputs of agent $i$, respectively.

\subsubsection{Separation Constraint}

To ensure the separation between agents, both hard and soft constraints are considered to address the accumulated error during prediction. That is, in the early stages, separation is enforced by a hard constraint, and after a predefined time \(T_{\rm sep} \in (0, T]\), it switches to a soft constraint. For a detailed justification, the reader is referred to \cite{aneek2}. Hard separation constraints are imposed as
\begin{equation}
\quad   d_{i,\text{sep}} - d_{j|i}^{t+k+1|t} \leq 0 , \quad k=0,...,T_{\text{sep}} -1,
\label{eq:sep_hard}
\end{equation}
where $d_{j|i}^{t+k+1|t} :=\| p_{i}^{t+k+1|t} -p_{j|i}^{t+k+1|t} \| ^{2}$ for \(j\in\overline{\mathcal{N}}^{t+k \mid t}_i\), and $d_{i,\text{sep}}$ is a separation distance. To add the soft constraint to the cost function, we define the penalty function as
\begin{equation}
\quad
    P( d_{i}^{t}) =\begin{cases}
    ( d_{i,\text{sep}} -d_{j|i}^{t})^{2} & \text{if} \ d_{i,\text{sep}}  >d_{j|i}^{t} ,\\
    0 & \text{otherwise}
    \end{cases},
\label{eq:sep_solf}
\end{equation}
then the objective function, equation (\ref{eq:cost}), will be modified as
\begin{equation}
    \tilde{J}(\textit{\textbf{u}}_{i}^{t}) =J(\textit{\textbf{u}}_{i}^{t}) + {\rho}_{\text{sep}}\sum _{k=T_{\text{sep}} +1}^{T -1} {\gamma} ^{k}\sum _{j\in \overline{\mathcal{N}}_{i}^{t+k|t}} P( d_{j|i}^{t+k|t}),
\label{eq:cost_star}
\end{equation}
where ${\rho}_{\text{sep}} \in \mathbb{R}_{+}$ acts as a penalty weight for a soft separation constraint. 

\subsection{Modified Flocking Rules} \label{modified_flocking}
This section will discuss the subjective neighbor weights and the weighted matrix $Q^{t}_{i}$. The effect of each rule was explained in the prior work (see \cite{aneek2}).
\subsubsection{Leader-follower graph-distance hierarchy}
A hierarchy level subjected to each agent ${\pi}^{t}_{i} \in \mathbb{N}$ was introduced. In the initialization process, each leader $l$ is set a constant hierarchy level ${\pi}^{t}_{l}=0$, and each follower is assigned an upper-bound hierarchy level $\bar{\pi} \in {\mathbb{N}}_{\geq 1}$, which can be defined as a flock parameter set by the designer. Then, the hierarchy level of each follower will be updated in every sample by the following equation
\begin{equation}
\quad \pi _{i}^{t} =\min\left\{\overline{\pi } ,1+\underset{j\in \overline{\mathcal{N}}_{i}^{t}}{\min} \pi _{j}^{t-1}\right\},
\label{eq:hierarchy}
\end{equation}
while the leaders' remain zero. In this way, each agent can estimate its current hierarchy level in a completely distributed manner. By communicating hierarchy levels between neighbors, each agent can identify the more important agents (e.g., leaders or followers near a leader). This information is incorporated into the position update rule by setting weight $w_{{\rm p},j|i}^{t+k|t}$ in equation (\ref{eq:y_bar}) as
\begin{equation}
\quad w_{{\rm p},j|i}^{t+k|t} =\frac{2^{-\pi _{j}}}{\sum _{\ell\in {\mathcal{N}}_{i}^{t+k|t}} 2^{-\pi _{\ell}}}.
\label{eq:w_p}
\end{equation}

\subsubsection{Allocating weights by travel vector}

The alignment weight should be prioritized for the agents in the front with respect to travel direction. This can be assessed by examining the inner product between its velocity and the relative position vector. The alignment weight $w_{{\rm v},j|i}^{t+k|t}$ can be defined as
\begin{equation}
\quad
    w_{{\rm v},j|i}^{t+k|t} =\begin{cases}
    1 & \text{if} \ \langle v_{i}^{t|t} ,p_{j|i}^{t-1|t-1} -p_{i}^{t|t} \rangle \geq 0\\
    \beta _{i} & \text{otherwise}
    \end{cases},
\label{eq:w_v}
\end{equation}
where ${\beta}_{i} \in [0,1].$

\subsubsection{Cohesion/alignment dynamic trade-off}
For each agent to have the ability to determine whether to prioritize cohering or aligning with its neighbor, the cohesion and alignment dynamic trade-off is implemented by dynamically adjusting the weight matrix $Q_{i}^{t}\in {\mathbb{R}}^{2n_{p} \times 2n_{p}}$ as follows
\begin{equation}
\quad Q_{i}^{t} :=\text{diag}\left( 1-q_{i}^{t} ,...,1-q_{i}^{t} ,q_{i}^{t} ,...,q_{i}^{t}\right)
\label{eq:Q}
\end{equation}
with
\begin{equation}
\quad q_{i}^{t} :=\frac{q_{i,\text{st}}}{1+c_{i} \| p_{i}^{t|t} -\overline{p}_{i}^{t|t} \| ^{2}},
\label{eq:q}
\end{equation}
where $c_{i}\in {\mathbb{R}}_{+}$, and $q_{i,\text{st}} \in (0,1)$. 

\section{Obstacle Avoidance Based on Local Sensor}
Point cloud data obtained from the depth camera or LiDAR sensor represents an environment around it in a robot's body frame, including obstacles and neighboring agents. For two-dimensional LiDAR, the number of data points can be large, depending on the sensor specification, and the number is squared for the three-dimensional one. In the optimization problem, it is not always possible to impose a high number of constraints due to the limitation of computational power. Hence, in this section, we present a point cloud processing technique for reducing the number of data points to a feasible range. Then, we define an obstacle avoidance constraint tailored to this processed data, aligning with our previously developed NMPC-based algorithm. The NMPC problem formulation with the proposed strategy is shown at the end of the section.


\subsection{Processing of Point Cloud Information}
In point cloud processing, we utilize directional filtering and down-sampling, which can significantly reduce the number of data points while preserving vital features of point cloud raw data. Since we do not consider the detected neighboring agents to be obstacles, the neighbor exclusion will be discussed in this section. The details for each process will be described in the following subsections.

We denote a set of processed point cloud's index perceived by agent $i$ at time $t$ that will be fed into the NMPC solver as ${\mathcal{O}}^{t}_{i}$. The symbol ${\mathcal{O}}^{t}_{\text{raw},i}$ refers to the index set of raw point cloud data from agent $i$'s sensor before any processing. Further, we define ${\mathcal{O}}^{t}_{{\rm f},i}$ and ${\mathcal{O}}^{t}_{{\rm s},i}$ as the index sets of the processed point cloud after directional filtering and down-sampling, respectively.

Since most of the sensors perceive the environment in their body frame, the point data is analyzed in the agent's body frame, where it is defined as follows: the x-axis points forward in the direction of its primary movement or orientation, representing its heading. The z-axis points upward, perpendicular to the plane of movement or operation. The y-axis completes the right-hand coordinate system, pointing to the robot's left side, orthogonal to the x-axis (forward) and the z-axis (up). Consequently, the robot's orientation can be briefly defined using the Euler angle ($\phi$, $\theta$, $\psi$), where ${\phi}^{t}_{i}$ denotes roll (rotation about the x-axis), ${\theta}^{t}_{i}$ signifies pitch (rotation about the y-axis), and ${\psi}^{t}_{i}$ represents yaw (rotation about the z-axis) of the $i^{th}$ agent at time $t$. Furthermore, vectors observed from the agents' body frame are denoted with the subscript ${\rm b}$.
\subsubsection{Directional Filtering}

Directional filtering is a step in the proposed point cloud processing to reduce the number of point cloud data points to be processed, allowing each agent to selectively process only the subset of environmental data relevant to its current motion intent. This strategy improves computational efficiency and robustness in dynamic and cluttered environments.

Each follower agent $i$ moves by minimizing a local objective function (\ref{eq:cost}) that encourages alignment with a desired direction of travel, represented by the weighted average position $\overline{p}{i}^{t|t}$ of its neighboring agents ${\mathcal{N}}{i}^{t}$. Intuitively, agent $i$ should prioritize environmental features that lie in the direction of this intended movement and disregard those in the opposite direction.

To formalize this directional prioritization, we define a reference plane in the body frame of agent $i$. The normal vector of this plane is given by
\begin{equation}
\quad
    \overline{p}_{{\rm b},i}^{t|t} =M\left( \phi _{i}^{t} ,\theta _{i}^{t} ,\psi _{i}^{t}\right) \cdot \left(\overline{p}_{i}^{t|t} -p_{i}^{t}\right),
\label{eq:p_bar_b}
\end{equation}
where $M\left( \phi _{i}^{t} ,\theta _{i}^{t} ,\psi _{i}^{t}\right)$ is a rotation matrix corresponding to the Euler angles of the $i^{th}$ agent with respect to the global frame. This transforms global frame vectors into agent $i$'s body frame. The reference plane passes through the origin of the body frame (i.e., the agent's center) and is orthogonal to $\overline{p}_{{\rm b},i}^{t|t}$. Data points in front of this plane (in the general direction of motion) are considered relevant.

Consider $p_{{\rm b},q|i}^{t} \in {\mathbb{R}}^{n_{p}}$ as the position vector of a data point detected by agent $i$'s sensor in its body frame, for $q \in {\mathcal{O}}^{t}_{\text{raw},i}$. Subsequently, we define a new subset ${\mathcal{O}}_{{\rm f},i}^{t} \subset {\mathcal{O}}_{\text{raw},i}^{t}$, representing the index set of point cloud data located on the positive side of a reference plane with normal vector $\overline{p}_{{\rm b},i}^{t|t}$. Accordingly, a data point $p_{{\rm b},q|i}^{t}$ belongs to ${\mathcal{O}}_{{\rm f},i}^{t}$ if and only if its inner product with $\overline{p}_{{\rm b},i}^{t|t}$ is positive. A filtered set of indices is defined as
\begin{equation}
\quad
    {\mathcal{O}}_{{\rm f},i}^{t} = \left\{q\in {\mathcal{O}}_{\text{raw},i}^{t} |\langle \overline{p}_{{\rm b},i}^{t|t} ,p_{{\rm b},q|i}^{t} \rangle \geq 0\right\},
\label{eq:df_O}
\end{equation}
which includes only the points located in front of the reference plane. This dot-product condition ensures that only the data aligned with or ahead of the movement direction are retained.

\begin{figure}[htbp]
\centering
\includegraphics[width=10cm]{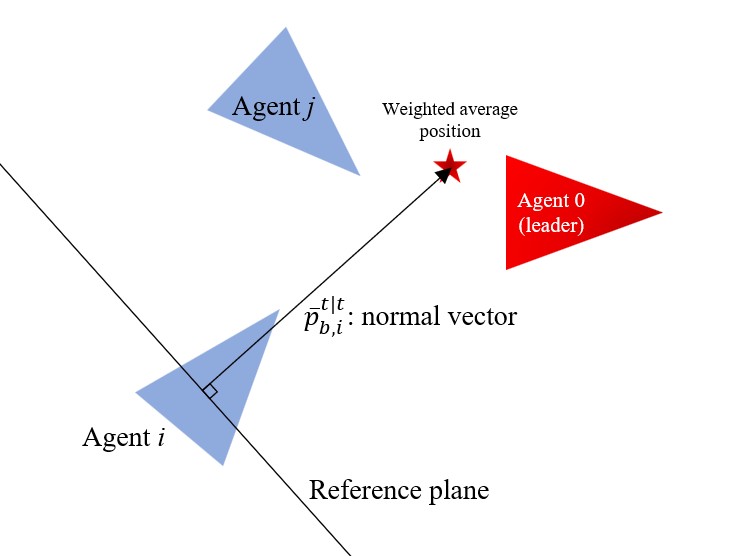}
\caption{\label{dir_1} The illustration of the reference plane in the Directional filtering process.}
\end{figure}

\begin{figure}[htbp]
\centering
\includegraphics[width=17cm]{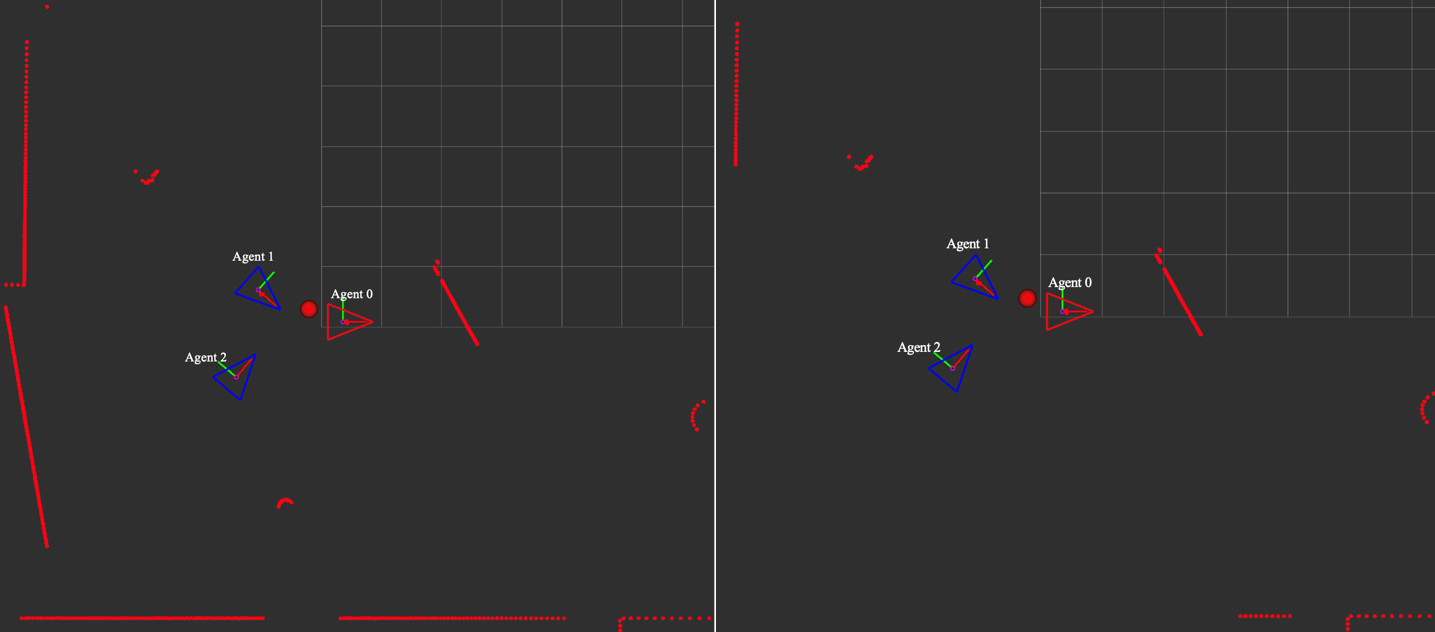}
\caption{\label{dir_2} Agent 2's raw point cloud visualization (left). Agent 2's point cloud visualization after employing Directional filtering (right). A red circle represents the weighted average among three robots, where the position of the leader agent (red triangle) is given more weight.}
\end{figure}

The illustration, portraying the reference plane and normal vector in the Directional filtering process, is displayed in Figure \ref{dir_1}, while the comparison of point cloud visualization in RViz between before and after is depicted in Figure \ref{dir_2}.

\subsubsection{Down-sampling}
In this process, data points are uniformly neglected by grouping nearby points and selecting the one closest to the robot from each group. Then, the indices of the selected points are assigned to a down-sampled point cloud index set ${\mathcal{O}}_{{\rm s},i}^{t} \subset {\mathcal{O}}_{{\rm f},i}^{t}$.

For two-dimensional point clouds, such as laser scan data, the index $q \in {\mathcal{O}}_{{\rm f},i}^{t}$ is organized based on spatial positioning. That is, the grouping process can be done by partitioning the polar coordinated laser scan data into sectors and selecting the point closest to agent $i$. A down-sampled index set can be represented as follows:
\begin{equation}
    {\mathcal{O}}_{{\rm s},i}^{t} = \Bigg\{ \underset{q}{\text{argmin}} 
    \Big\{ r_{q} \mid q\in {\mathcal{O}}_{{\rm f},i}^{t}[ i_{s}:i_{s}+f_{s}] \Big\}  \mid i_{s}\in \{0,f_{s},2f_{s},\ldots,n_{s}f_{s}\} \Bigg\},
\label{eq:ds_O_2d}
\end{equation}
where $f_{s} \in \mathbb{R}$ is the down-sample factor and $n_{s}=\lfloor ( |{\mathcal{O}}^{t}_{{\rm f},i} |-1) /f_{s}\rfloor$. The index set ${\mathcal{O}}_{{\rm f},i}^{t}[ i_{s}:i_{s}+f_{s}]$ denotes the segment for $f_{s}$ consecutive index elements in ${\mathcal{O}}_{{\rm f},i}^{t}$ starting at index $i_{s}$, and ending with $i_{s}+f_{s}$. The distance $r_{q}$ ranges from the center of agent $i$ to the $q^{th}$ point data.

For three-dimensional point clouds, where the data points can be gathered from a depth camera or 3D LiDAR, the perceived data is unstructured, unlike in the two-dimensional case. A widely used method for down-sampling such data is voxel grid filtering proposed by \cite{vgf1}, \cite{vgf2}. In this work, a modified voxel grid filtering technique will be utilized by prioritizing the closest point in each voxel. The algorithm creates voxel volume grids over the point cloud data, and the down-sampled point for each grid will be the one closest to the $i^{th}$ agent. Let $p_{{\rm b},q|i}^{t} =\left[ x_{{\rm b},q|i}^{t} ,y_{{\rm b},q|i}^{t} ,z_{{\rm b},q|i}^{t}\right]^{\top}$ be a position of the obstacle $q \in {\mathcal{O}}_{{\rm f},i}^{t}$ in three-dimensional space observed from agent $i$'s body frame, and $V:  {\mathbb{N}} \rightarrow  {\mathbb{Z}}^{3}$ be a function that maps the obstacle's index to its corresponding voxel coordinates in a three-dimensional grid space defined as
\begin{equation}
    V( q) =\left[\left\lfloor \frac{x_{{\rm b},q|i}^{t}}{\delta _{x}}\right\rfloor ,\left\lfloor \frac{y_{{\rm b},q|i}^{t}}{\delta _{y}}\right\rfloor ,\left\lfloor \frac{z_{{\rm b},q|i}^{t}}{\delta _{z}}\right\rfloor \right]^{\top},
\label{eq:v}
\end{equation}
where $\delta_{x}$, $\delta_{y}$, and $\delta_{z}$ denote voxel grid size. The set expression can be represented as
\begin{eqnarray}
{\mathcal{O}}_{{\rm s},i}^{t} &= \Bigg\{q\in {\mathcal{O}}_{{\rm f},i}^{t} \, \Big| \, \nexists q' \in {\mathcal{O}}_{{\rm f},i}^{t}: \, (V(q) = V(q')) \wedge \nonumber 
\, (q \neq q') \wedge ( \, \| p_{{\rm b},q' |i}^{t} \| ^{2} < \| p_{{\rm b},q|i}^{t} \| ^{2} ) \Bigg\}.
\end{eqnarray}
According to the above equation, the set ${\mathcal{O}}_{{\rm s},i}^{t}$ consists of the indices $q$ for which there is no other index $q'$ in the set ${\mathcal{O}}_{{\rm f},i}^{t}$ (after the directional filtering process) such that the obstacle corresponding to index $q'$ is in the same voxel as obstacle $q$, and is closer to the agent $i$.

\subsubsection{Neighbor exclusion}
The point cloud data perceived via a sensor contains obstacles and neighboring agents. The neighbors could be considered dynamic obstacles, as in \cite{bjorn1}. However, in this work, the imposed separation rule was highlighted. Thus, the detected neighbors in the point cloud index can be neglected to reduce the number of data points.

This is the final stage of the point cloud processing, where the set of processed data ${\mathcal{O}}^{t}_{i}$ will be obtained and then used in the optimal control problem. The set expression of the processed point cloud is expressed as follows:
\begin{equation}
\thinspace
{\mathcal{O}}_{i}^{t} =\left\{q\in {\mathcal{O}}_{{\rm s},i}^{t} |\| p_{q|i}^{t} -p_{j|i}^{t|t} \|  >r_{b} ,j\in \overline{\mathcal{N}}_{i}^{t|t}\right\},
\label{eq:processed_O}
\end{equation}
where $r_{b}$ is the farthest distance within the agent's body from its center.

\subsection{Obstacle Avoidance Constraint}
To ensure obstacle-free trajectories along the finite prediction horizon for each agent, we impose an obstacle-avoidance constraint. Since obstacles are defined as processed point clouds in $n_{p}$-dimensional space, let $p^{t}_{m|i} \in {\mathbb{R}}^{n_{p}}$ be an $m^{th}$ obstacle's position sensed by the $i^{th}$ agent, where $m \in {\mathcal{O}}^{t}_{i}$, and $r_{s} \in {\mathbb{R}}_{+}$ be a safety distance away from an obstacle. The obstacle can be expressed as
\begin{equation}
\quad
    h(p_{i}^{t},p_{m|i}^{t}) = r_{s}^{2} -\| p_{i}^{t} -p_{m|i}^{t} \| ^{2}.
\label{eq:obs_def}
\end{equation}
The obstacle avoidance condition is that the trajectory of an agent $i$ along the NMPC's prediction horizon has to be completely outside the sphere or circle defined by equation (\ref{eq:obs_def}). Thus, the obstacle avoidance constraint can be defined as: 
\begin{equation}
\quad
    h(p_{i}^{t+k+1|t},p_{m|i}^{t}) \leq 0, \quad k=0,..., T-1.
\label{eq:obs_constraint}
\end{equation}

\subsection{NMPC Problem}
From the objective function (\ref{eq:cost_star}) and the constraints (\ref{eq:state_input_constraint}), (\ref{eq:sep_hard}), and (\ref{eq:obs_constraint}) described earlier, the optimization problem can be formulated  as
\begin{eqnarray}
\begin{array}{rcl}
\underset{\textit{\textbf{u}}_{i}^{t}}{\text{Minimize}} \thinspace \thinspace \tilde{J}(\textit{\textbf{u}}_{i}^{t})
\end{array}
\label{eq:cost_mpc_problem}
\end{eqnarray}
subject to:
\begin{align}
\begin{array}{l}
\left.\begin{aligned}
&x_{i}^{t+k+1|t} =f_{i} (x_{i}^{t+k|t} ,u_{i}^{t+k|t} ) \\
&u_i^{t+k|t} \in {\mathcal{U}}_i \\
&x_i^{t+k+1|t} \in {\mathcal{X}}_i \\
&h(p_{i}^{t+k+1|t},p_{m|i}^{t}) \leq 0,\thinspace m \in {\mathcal{O}}_i^t \\
\end{aligned} \right\} \text{for } k = 0,..., T-1 \\
\begin{aligned}
&d_{j|i}^{t+k+1|t} \geq d_{i,\text{sep}}, \thinspace j \in \overline{{\mathcal{N}}}_{i}^{t+k+1|t} \thinspace \text{for } k = 0,..., T_{\text{sep}}-1
\end{aligned}
\end{array}
\end{align}
This problem will be fed into the OpEn (see \cite{OpEn}) framework to generate Rust code that solves the constrained optimization problem using PANOC (see \cite{PANOC}) with augmented Lagrangian and penalty methods.

\section{Simulation}
\subsection{Simulation Setup}
In this work, the proposed optimal control problem is demonstrated through a close-to-real-world simulation in Gazebo (see \cite{gazebo}), which incorporates physical quantities and three-dimensional dynamics, e.g., friction and inertia. This simulation environment is well integrated with ROS (Robot Operating System). The agent model in our study is the unicycle ground vehicle model, chosen for its representation of most mobile robots and its inherent nonlinear properties as a nonlinear input affine system.

The continuous-time state equation of the unicycle ground vehicle model is given by
\begin{equation}
\quad
\begin{bmatrix}
\dot{p}_{i,x}( t)\\
\dot{p}_{i,y}( t)\\
\dot{\psi }_{i}( t)
\end{bmatrix} =\begin{bmatrix}
\cos \psi _{i}( t) & 0\\
\sin \psi _{i}( t) & 0\\
0 & 1
\end{bmatrix}\begin{bmatrix}
v_{i}( t)\\
\dot{\psi }_{i}( t)
\end{bmatrix},
\end{equation}
where $v_{i}(t)$ and $\dot{\psi}_{i}(t)$ are the system inputs. The discrete-time state update equations can be derived through a forward Euler discretization. Additionally, the extra states, which are translational velocities with respect to the global frame, are added so that the alignment with its neighbors will be explicitly determined. Hence, the system's state becomes $x_{i}^{k} =\left[ p_{i,x}^{k} ,p_{i,y}^{k} ,\psi _{i}^{k} ,v_{i,x}^{k} ,v_{i,y}^{k}\right]^{\top} \in {\mathbb{R}}^{5}$, and the system's input is $u_{i}^{k} =\left[ v_{i}^{k} ,\dot{\psi }_{i}^{k}\right]^{\top} \in {\mathbb{R}}^{2}$. The discrete-time state equation can be expressed as
\begin{equation} 
\quad \begin{bmatrix}
p_{i,x}^{k+1}\\
p_{i,y}^{k+1}\\
\psi _{i}^{k+1}\\
v_{i,x}^{k+1}\\
v_{i,y}^{k+1}
\end{bmatrix} =\begin{bmatrix}
p_{i,x}^{k}\\
p_{i,y}^{k}\\
\psi _{i}^{k}\\
0\\
0
\end{bmatrix} +\begin{bmatrix}
\Delta t\cos \psi _{i}^{k} & 0\\
\Delta t\sin \psi _{i}^{k} & 0\\
0 & \Delta t\\
\cos \psi _{i}^{k} & 0\\
\sin \psi _{i}^{k} & 0
\end{bmatrix}\begin{bmatrix}
v_{i}^{k}\\
\dot{\psi }_{i}^{k}
\end{bmatrix},
\label{kinematic}
\end{equation}
where $\Delta t$ is the sampling time in seconds. This discretized state equation will be used in NMPC to update the agent's trajectory along the finite horizon period. 

\begin{figure}[htbp]
\centering
\includegraphics[width=15cm]{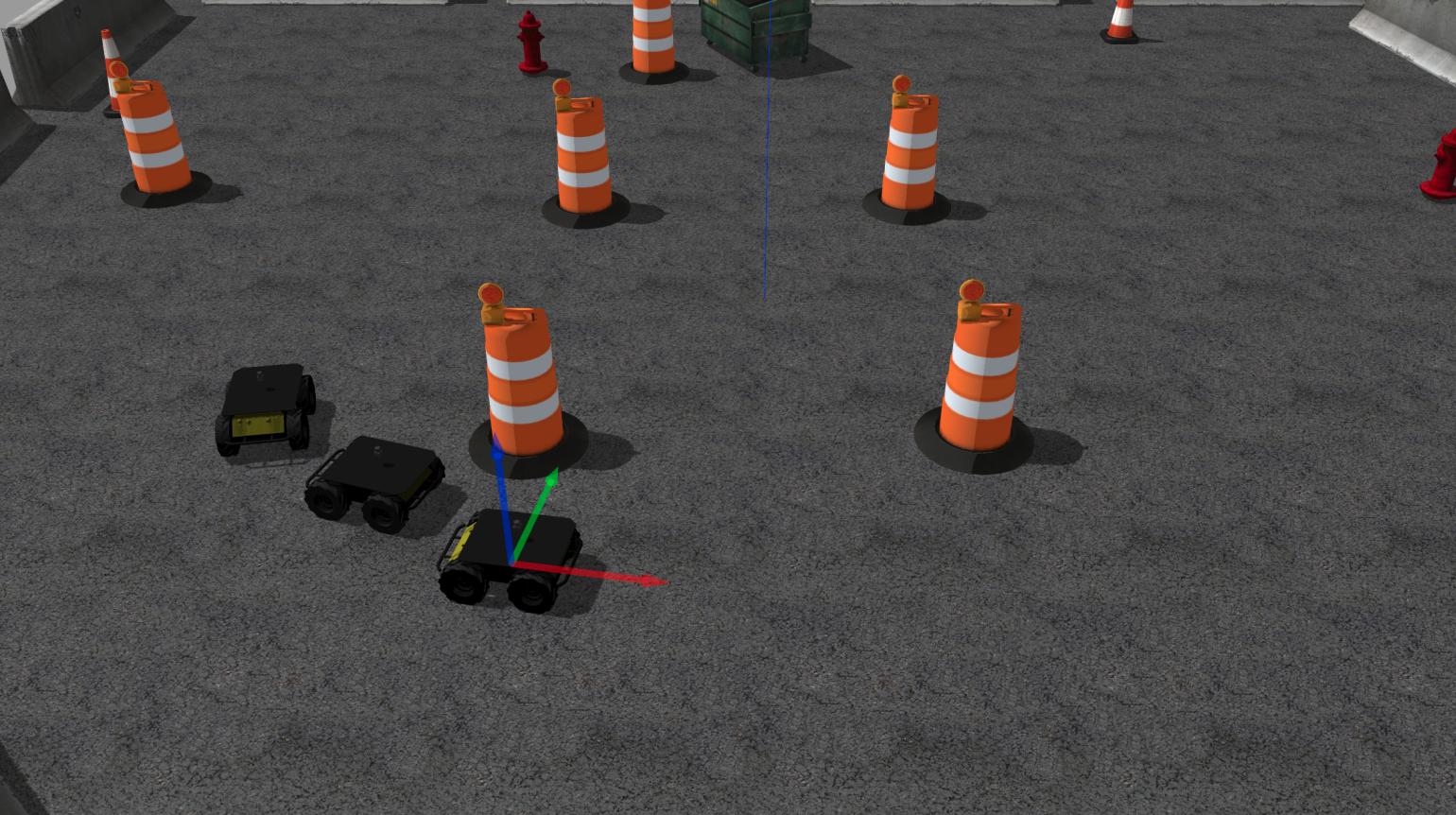}
\caption{\label{gazebo} The Gazebo 3D simulation of three Husky UGVs from \cite{husky} in an obstructed environment by \cite{playen}.}
\end{figure}

\begin{figure}[htbp]
\centering
\includegraphics[width=15cm]{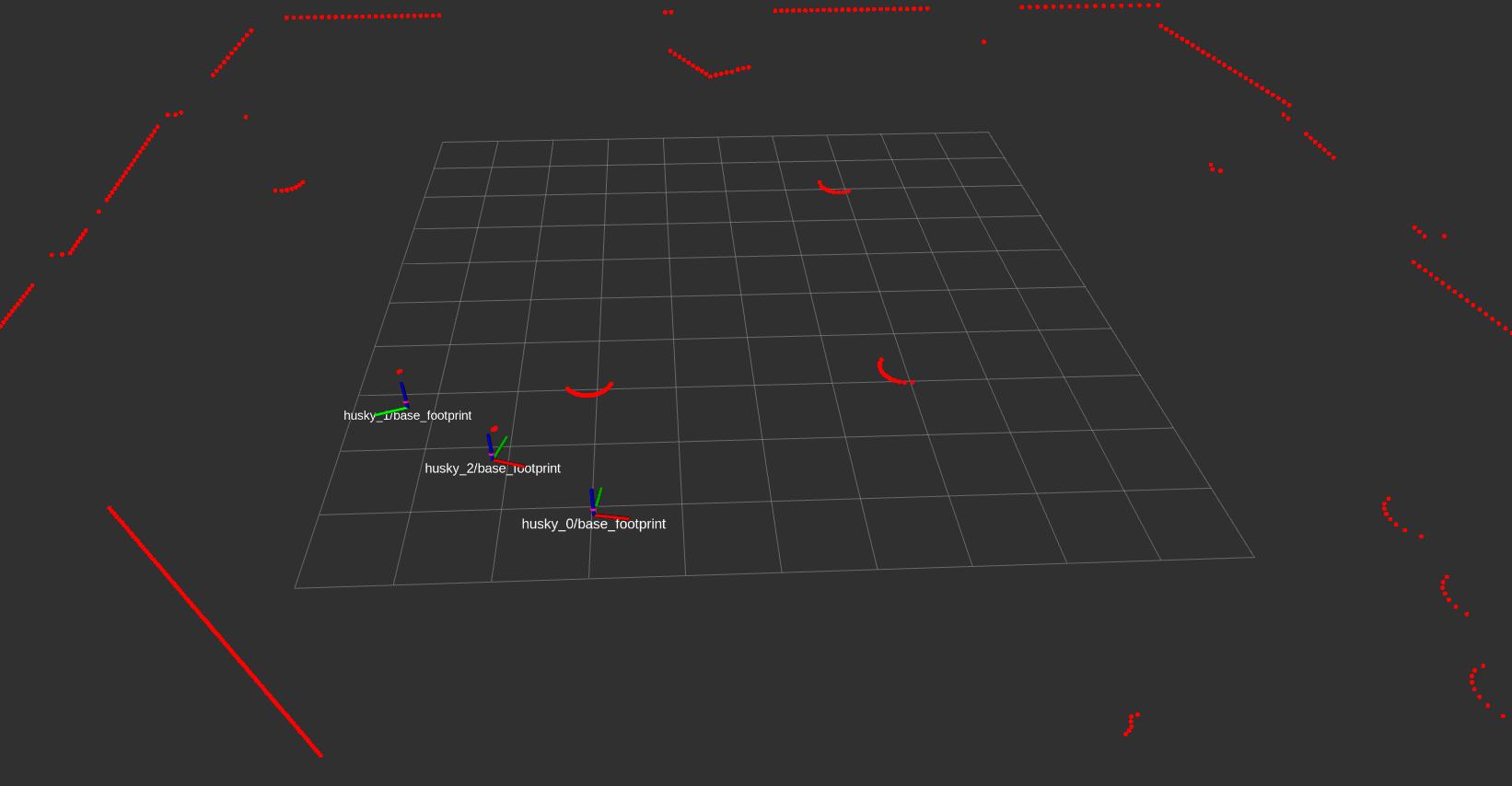}
\caption{\label{laser} The visualization of laser scan point cloud data from UST10 simulated 2D LiDAR and the robots' frames in RViz.}
\end{figure}

We utilized the Husky UGV from \cite{husky} as agents in the 3D Gazebo simulation. Figure \ref{gazebo} displays the simulated Husky UGVs and environment in our Gazebo 3D simulation. The robots are attached with a UST10 simulated 2D LiDAR at their body center. The LiDAR has a default range of 5 meters, 360 degrees angular range starting from the robot heading, and rotates counterclockwise, with 720 sampling points per round in the robot's body frame polar coordinate. The output information from LiDAR contains a time stamp, frames, sensor configurations, and ranges of detected obstacles packed in the laser scanner message type in the ROS convention, which are two-dimensional point clouds. These raw data will be fed into the point cloud processor to produce the processed point data and then provided to the NMPC solver. The demonstration of the point cloud data is shown through RViz (ROS visualization) in Figure \ref{laser}. The video of the 3D simulation with visualized point cloud data can be accessed via the following link \textbf{https://youtu.be/APg52Rw725M}.

Note that the state equation (\ref{kinematic}) represents the kinematic model of a unicycle mobile robot, which omits physical characteristics such as the relationship between motor voltage, current, torque, and speed. In other words, this model serves as a simplified representation for the proposed NMPC, where the controller generates high-level commands in the form of translational and angular velocities. In a real-world implementation, these commands would typically be passed to a low-level controller, such as a PID controller, which regulates motor actuation by adjusting electrical inputs accordingly. In this study, the high-level commands are sent to a virtual low-level system that converts the unicycle model's control inputs into individual wheel speeds and torques. In the simulation, the wheels' speeds are sensed, and torques are regulated using well-tuned low-level PID controllers. A complete model incorporating other dynamic characteristics could be considered for more accurate state prediction in an NMPC scheme. However, it requires system identification techniques and is not in our scope.

\begin{figure}[htbp]
\centering
\includegraphics[width=15cm]{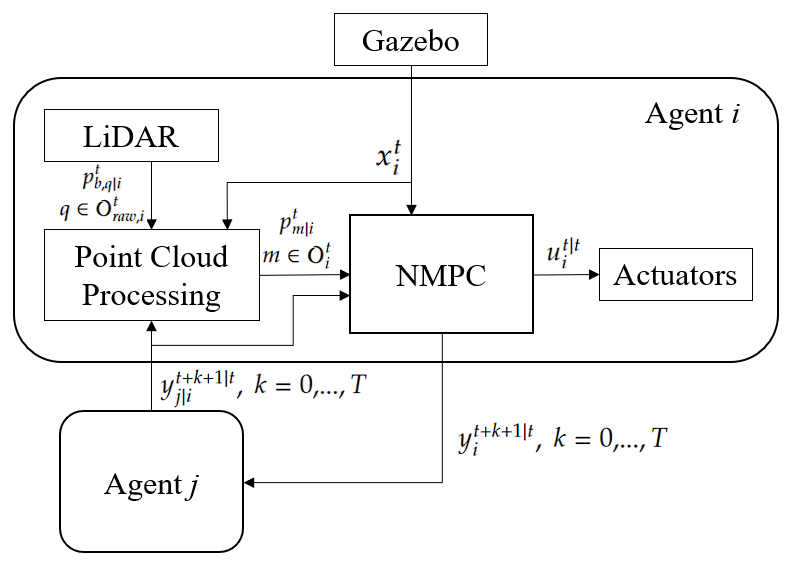}
\caption{\label{system} The schematic diagram of an agent $i$'s distributed NMPC with point cloud processing.}
\end{figure}

In the simulation, we consider the system with three agents, consisting of one leader and two followers. The leader will accept the arbitrary input command and predict its translational position and velocity trajectory within the horizon period $T$ based on the instantaneous inputs. The predicted trajectory will be shared with the followers within the detection range, which is assumed to be the same as the LiDAR default range. The followers also share their predicted trajectories from the NMPC optimizer with one another. The schematic diagram of follower agent $i$ is shown in Figure \ref{system}, where agent $j$ is the neighbor within agent $i$'s detection range ($j \in \overline{\mathcal{N}}_{i}^{t+k|t}$).

The MPC horizon for all agents is set to $T=10$ steps. The separation horizons are set as $T_{\text{sep}}=5$ steps, and the prediction of collision avoidance beyond $T_{\text{sep}}$ is softly constrained with a penalty parameter of ${\rho}_{\text{sep}}=20$. The discount factor $\gamma$ is set to $0.8$. For the alignment weights of agents positioned behind, as stated in equation (\ref{eq:w_v}), a value of ${\beta}_{i}=0.5$ is adopted. In the trade-off rule described in equation (\ref{eq:q}), a default static coefficient of $q_{i,\text{st}}=0.5$ is utilized alongside a dynamic weight of $c_{i}=10$. An upper limit of $\overline{\pi}=3$ is set to be equal to the total number of agents (including the leader) for the hierarchy levels according to equation (\ref{eq:hierarchy}). The sampling time for state equation discretization $\Delta t$ is $0.1$ second. Since we employ a two-dimensional laser scanner as a sensor in our simulation, the down-sampling processing is utilized based on equation (\ref{eq:ds_O_2d}), where the down-sampling factor $f_{s}$ is set to 4.

We employed rectangular sets for the inputs and state constraints. The input boundaries for translational velocity are $[-0.1, 1.0]$ m/s and $[-8,8]$ rad/s for rotational velocity. The state constraint can be interpreted as a square workspace in a two-dimensional plane originating at $(0,0)$, in which the robots can move regardless of obstacles. We define the spatial limit as $p_{x} =p_{y}=\pm 10$ meters. For obstacle avoidance constraint, the inequality in (\ref{eq:obs_constraint}) is reformulated to $\min \{ 0, h(p_{i}^{t+k+1|t},p_{m|i}^{t}) \}=0$, for $k=0,...,T-1$, and $m \in {\mathcal{O}}^{t}_{i}$. The obstacle avoidance constraint is implemented through OpEn's penalty method, while the state and separation (for $k < T_{\text{sep}}$) constraints are handled using the augmented Lagrangian method. The solver parameters are configured as follows: the solution tolerance $\epsilon$ and $\delta$ are set to $1 \times 10^{-5}$ and $1 \times 10^{-4}$ respectively. The initial penalty ${\lambda}_{0}$ is set to $1 \times 10^{-2}$. The penalty update coefficient $\rho$, which is different from the separation soft constraint coefficient, is selected to be $5$.

\subsection{Navigate in the Obstructed Environment}



\begin{figure}[htbp]
\centering
\includegraphics[width=10cm]{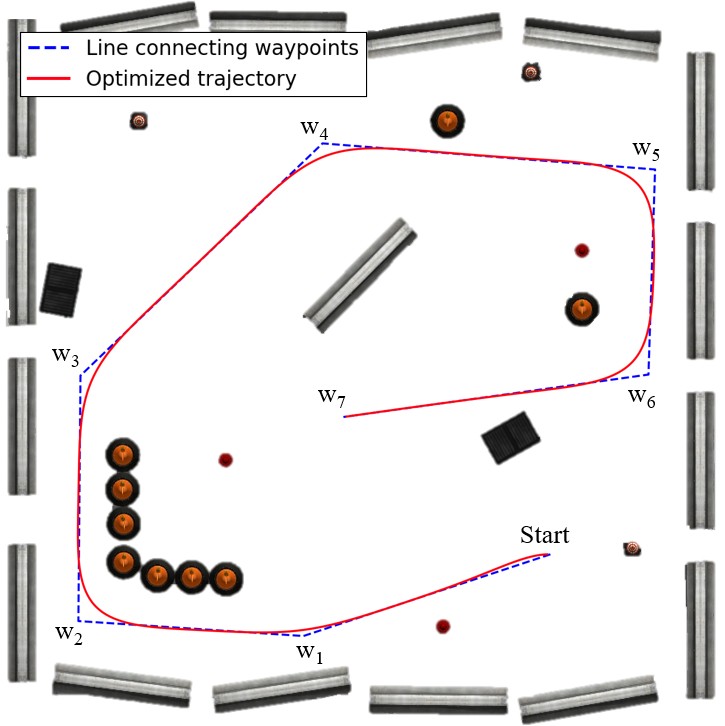}
\caption{\label{opt_traj} The reference trajectory is shown in red, and the lines connecting waypoints are shown in broken blue. $\text{w}_{i}$ for $i=1, \dotsc 7$ are waypoints.}
\end{figure}

We tested the proposed optimal control formulation and algorithm by letting the leader agent track a reference trajectory in an obstructed environment created by \cite{playen} and evaluating the followers' trajectories resulting from the NMPC problem formulated in section 3.3. The environment contains various obstacles, such as barriers, fire hydrants, dumpsters, and construction cones. The reference trajectory can be obtained by first specifying waypoints. Then, create points between the two immediate waypoints, where the distances between points are constant except for the remaining segments. Suppose that $T_{p}$ is the total number of points, and $\hat{P} = \{ \hat{p}^{0},\dotsc, \hat{p}^{T_p} \}$ is a set of points from the given process, the reference trajectory for the leader agent $\prescript{\text{ref}}{}{} P= \{ p^0 , \dotsc, p^{T_p} \}$ is defined as
\begin{equation} \label{opt_traj_eq}
    \prescript{\text{ref}}{}{} P =\underset{p^{k} ,u^{k}}{\arg\min}\sum _{k=0}^{T_{p} -1}\left( q_{p} \| p^{k} -\hat{p}^{k} \| ^{2} +q_{u} \| u^{k} \| ^{2}\right) +q_{T} \| p^{T_{p}} -\hat{p}^{T_{p}} \| ^{2}, \quad k=0,\dotsc,T_p,
\end{equation}
subjected to the state equation (\ref{kinematic}), where $q_p \in \mathbb{R}$, $q_T \in \mathbb{R}$ and $q_u \in \mathbb{R}$ are weighted constants. The optimized trajectory, a so-called admissible path for unicycle mobile robots, is depicted in Figure \ref{opt_traj}. The leader agent will track this reference trajectory in (\ref{opt_traj_eq}) employing a simple reactive controller, and the followers will preserve the fleet connectivity using the proposed NMPC scheme. Let's say the agent index for the leader is $0$. Its control law is defined below.
\begin{equation} \label{reactive_control}
    \begin{aligned}
        v_{0}^{k} &=K_{v} \| \hat{p}^{k} -p^{k}_{0} \| ^{2} \\
        \dot{\psi }_{0}^{k} &=K_{\psi } \cdot \text{wrap}\left(\hat{\psi }^{k}_{0} -\psi ^{k}_{0}\right), \quad k = 0, \dotsc , T_p,
    \end{aligned}
\end{equation}
where $K_v \in \mathbb{R}$ and $K_\psi \in \mathbb{R}$ are the proportional gains for each control channel. $\hat{\psi }^{k}$ is defined as

\begin{equation}
    \hat{\psi }^{k}_{0} =\arctan\left[\frac{\left(\hat{p}^{k} -p^{k}_{0}\right)_{y}}{\left(\hat{p}^{k} -p^{k}_{0}\right)_{x}}\right].
\end{equation}
A function $\text{wrap}:\mathbb{R}\rightarrow \mathbb{R}$ is a function that wraps the angle to be within the range of $[0, \pi )$, defined as
\begin{equation}
    \text{wrap}(\theta) = \theta - 2\pi \cdot \left\lfloor \frac{\theta}{2\pi} \right\rfloor.
\end{equation}

\begin{figure}[htbp]
\centering
\includegraphics[width=10cm]{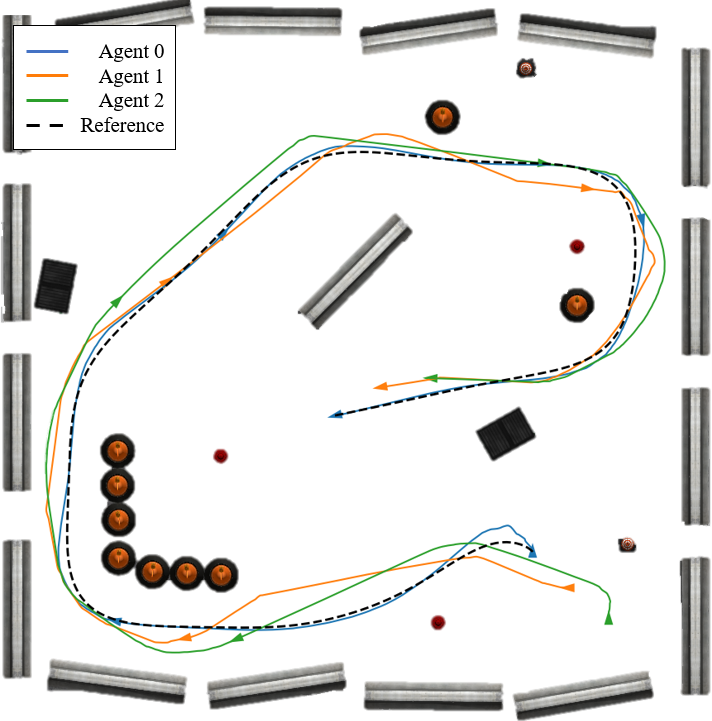}
\caption{\label{traj} The trajectories of 3 agents in an obstructed environment, where the blue trajectory is the leader's and the rest are followers.}
\end{figure}

\begin{figure}[htbp]
\centering
\includegraphics[width=13cm]{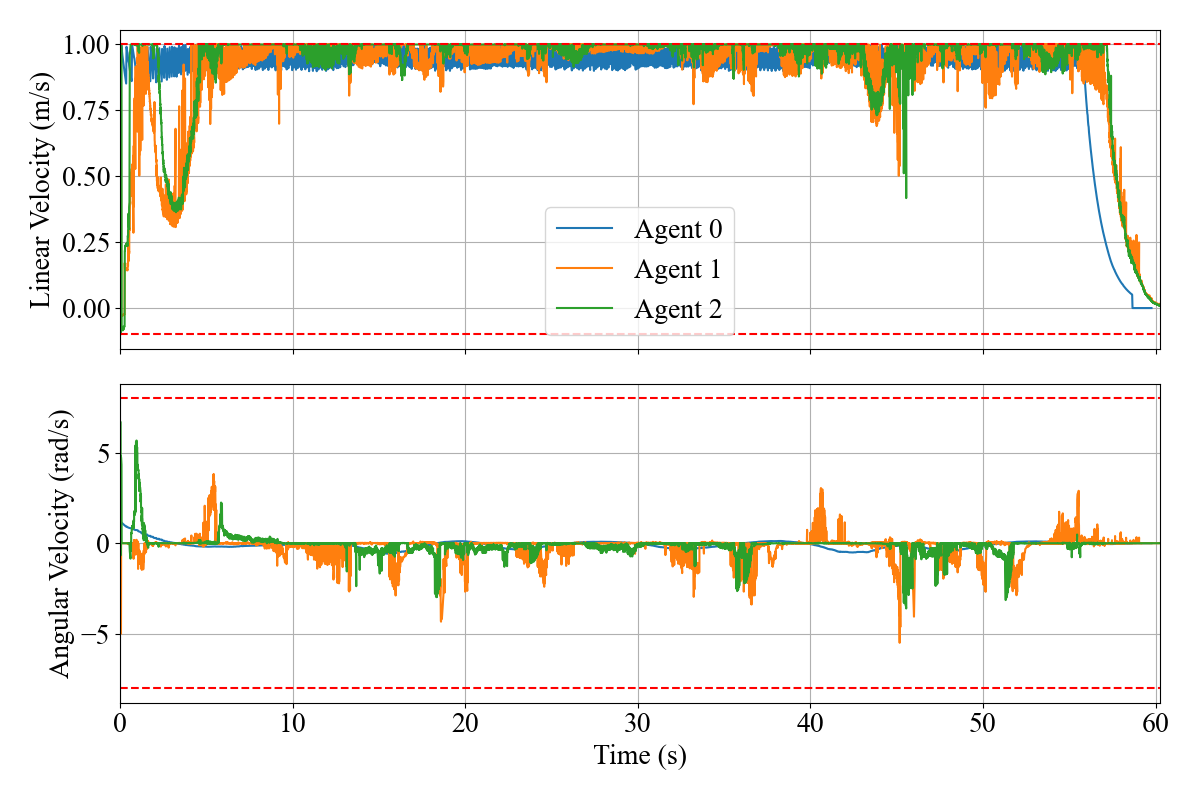}
\caption{\label{input_pc} Time-series control inputs with constraint boundary, shown as broken red lines.}
\end{figure}

The initial poses of the robots are: Agent 0 at (5.86, -5.13) with 0 radians (the same pose as the initial waypoint in Figure \ref{opt_traj}), Agent 1 at (6.85, -6.25) with 1.5 radians, and Agent 2 at (7.87, -7.35) with 0 radians, where Agent 0 is the leader, and the rest are followers. The resulting trajectories are shown in Figure \ref{traj} with the reference trajectory illustrated as a broken black path, and the control inputs throughout the experimentation are displayed in Figure \ref{input_pc} with the control inputs' boundary shown as broken red lines. The simulation video can be accessed through the following link \textbf{https://youtu.be/u8TcpLy7NKI}. In the simulation, the leader is given a control input (\ref{reactive_control}) to track the optimized path, and the two followers can maintain fleet connectivity and separation while avoiding obstacles using local information. In the demo, each follower recognized up to two neighbors if they are in the sensor's range, regardless of the point cloud processing, which only simplifies the obstacle avoidance constraint. The algorithm is scalable, but we only showed three robots for visualization purposes.

\begin{figure}[htbp]
\centering
\includegraphics[width=15cm]{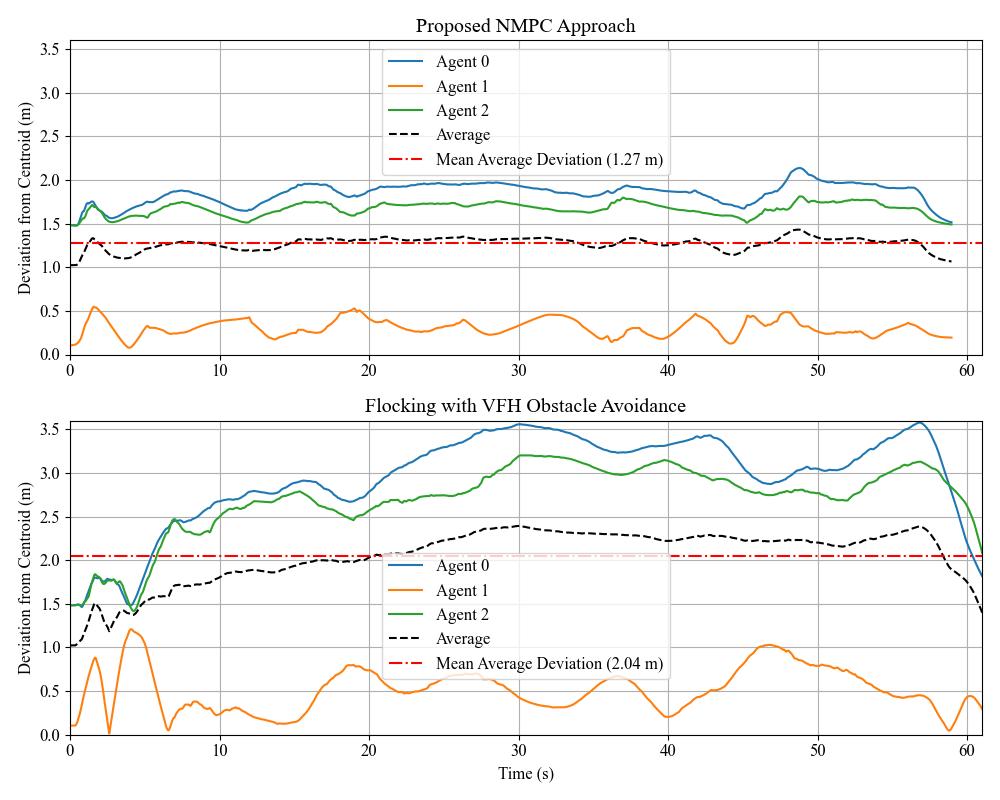}
\caption{\label{vs_vfh} The plot of the deviation from the centroid of the two approaches.}
\end{figure}

Furthermore, a comparison experiment is conducted to benchmark the proposed NMPC algorithm. For this purpose, the vector field histogram (VFH) algorithm, tailored for flock navigation, is employed. VFH is a simple reactive obstacle avoidance strategy well-suited for a two-dimensional LiDAR sensor, where obstacle information is stored in polar coordinate form. The algorithm converts obstacle data into a one-dimensional histogram and selects a sector that is both obstacle-free and closest to the target, allowing the robot to navigate accordingly. See \cite{vfh} for more details on this algorithm. The simulation is performed in a similar manner, where the leader agent tracks the reference path (\ref{opt_traj_eq}), and the followers execute the tailored VFH algorithm to maintain fleet connectivity. The fleet connectivity is quantified by the deviation from the average position of all agents, i.e., the deviation from the centroid. The plot comparing the deviation from the centroid for both the proposed method and the VFH algorithm is shown in Figure \ref{vs_vfh}. The results indicate that the proposed algorithm navigates the robot fleet through an obstructed environment with better connectivity preservation.

\subsection{Hardware-in-the-Loop Simulation}
\begin{figure}[htbp]
\centering
\includegraphics[width=10cm]{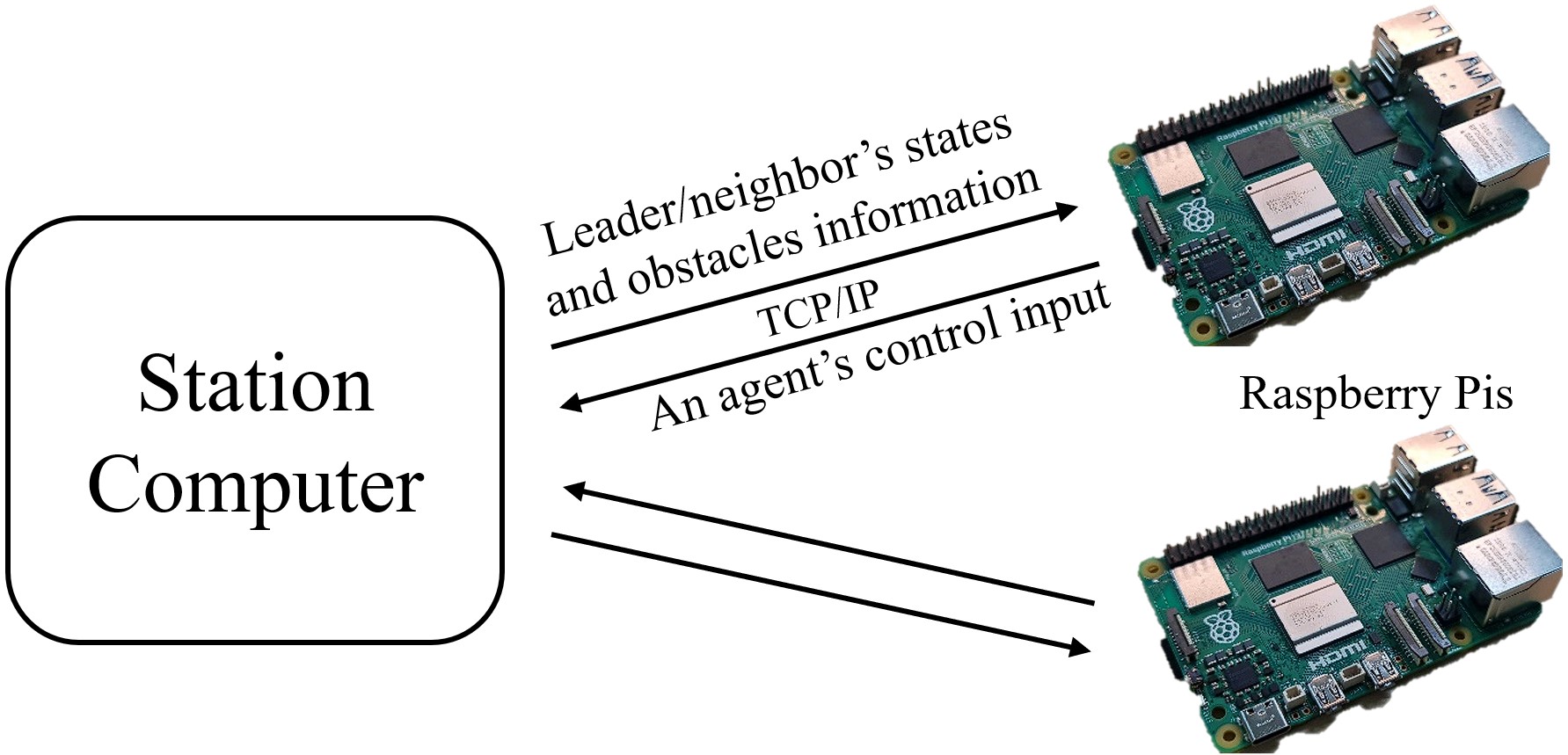}
\caption{\label{hil_scheme} Schematic diagram of the HIL experimentation: The algorithm for the two followers is executed in each Raspberry Pi, while the station computer is responsible for the entire simulation. The information is shared among them using TCP/IP-based client-server communication protocol with ROS.}
\end{figure}

In the previous subsection, the optimal control problem was solved successfully in a distributed manner on a station computer. However, the embedded processor might not have enough computational power to solve such a problem smoothly. Thus, in this section, we examine the feasibility of implementing the proposed NMPC on Raspberry Pi by connecting it to a station computer and performing a hardware-in-the-loop (HIL) simulation. In this setting, we use two Raspberry Pis to represent the two followers, Agents 1 and 2. The board model we use is Raspberry Pi 4 B, with a 1.5 GHz 64-bit quad-core ARM Cortex-A72 processor (ARM architecture), 1500 MHz clock speed, and 4GB of RAM. The HIL simulation is conducted through a master-slave communication scheme, where the station computer is assigned to be a master in a ROS fashion. We connect the two Raspberry Pis to the same network as the simulation computer and create the communication bridge. Then, ROS topics published by either the master or slaves can be subscribed to by every machine within this network. The NMPC solvers operate independently on each target board while exchanging information. The time required for the optimizer to attain the solution is crucial. The longer it takes, the more performance degradation. To ensure that the solution for each optimizer is available within the discretization sampling period, said 100 ms, we set the solve time cut-off to 95 ms, meaning that the solver will stop the operation once the solving time reaches the specified cut-off, and the lowest cost value in the last iteration will be the result. The leader is given a reference trajectory to track, akin to the simulation in the preceding section, utilizing the robots' initial poses from the previous section. The schematic diagram of the experimentation is depicted in Figure \ref{hil_scheme}.

\begin{figure}[htbp]
\centering
\includegraphics[width=10cm]{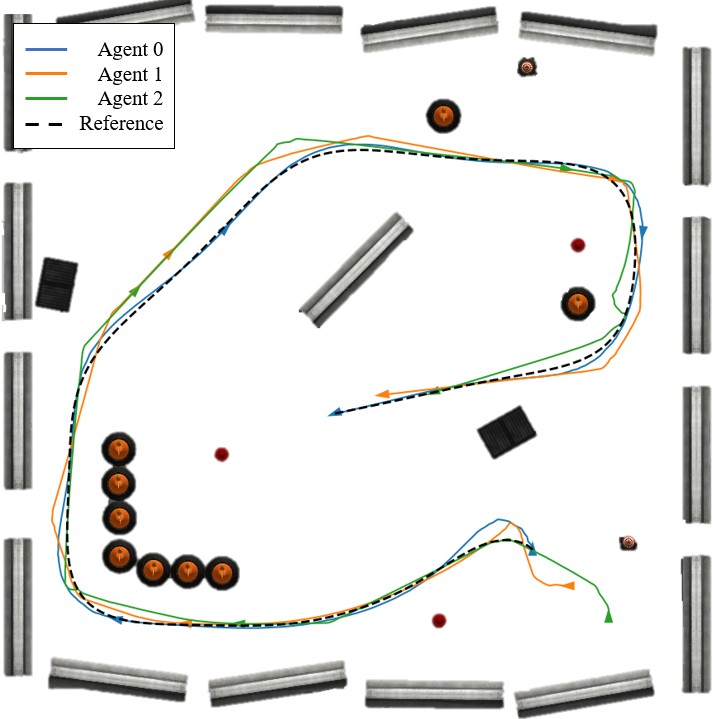}
\caption{\label{rpi_traj} HIL simulated trajectories of 3 agents in an obstructed environment, where the blue trajectory is the leader's and the rest are followers.}
\end{figure}

\begin{figure}[htbp]
\centering
\includegraphics[width=13cm]{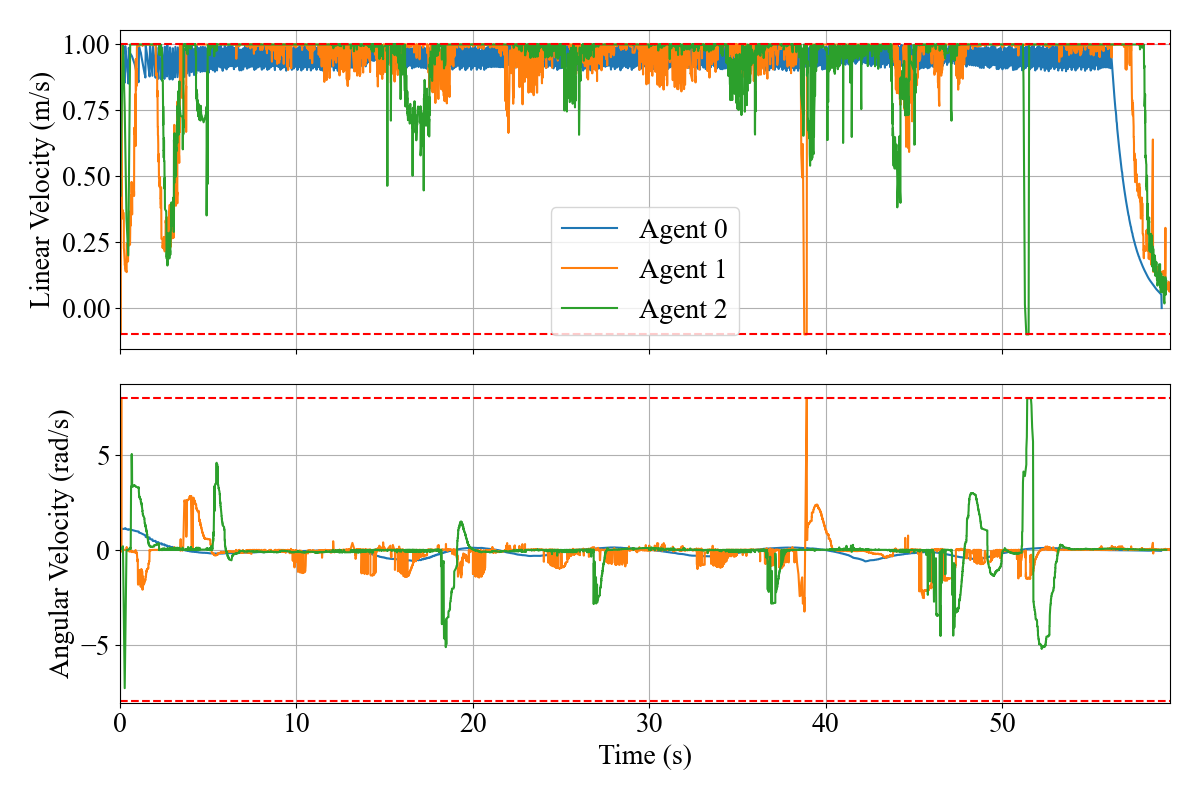}
\caption{\label{input_rpi} HIL simulation's time-series of the control inputs with boundary, depicted in broken red.}
\end{figure}

\begin{figure}[htbp]
\centering
\includegraphics[width=13cm]{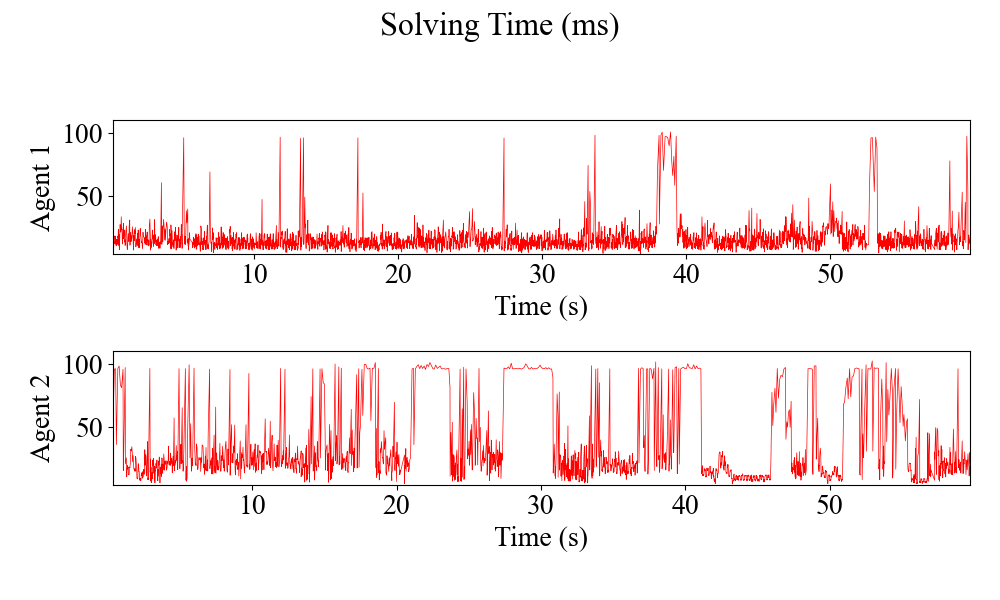}
\caption{\label{rpi_solve_time} NMPC's solver solving time on Raspberry Pi 4B: agent 1 averages 17.19 ms and peaks at 95 ms, while agent 2 averages 24.6 ms and also peaks at 95 ms.}
\end{figure}

\begin{figure}[htbp]
\centering
\includegraphics[width=13cm]{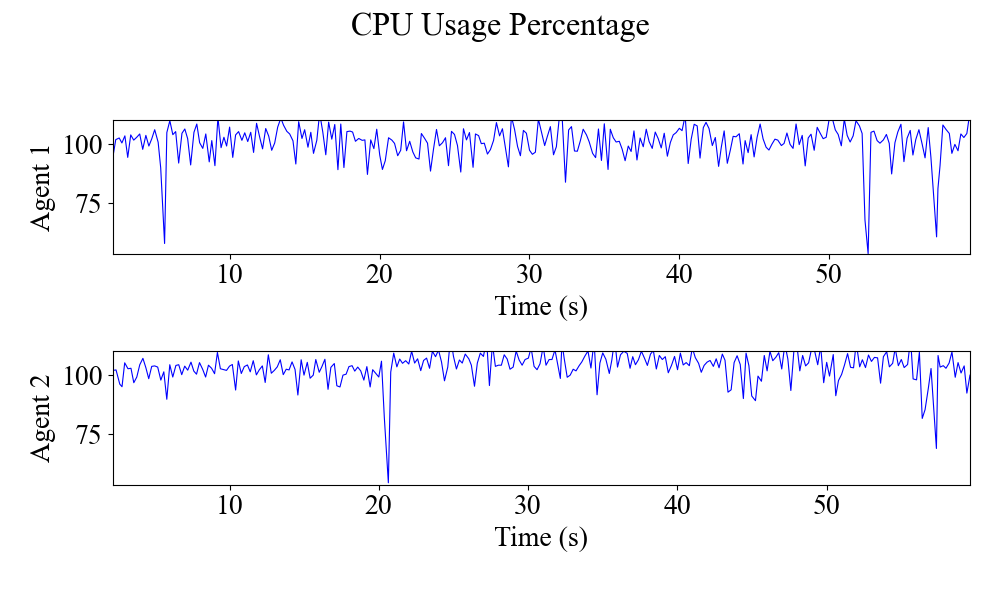}
\caption{\label{rpi_cpu} The Raspberry Pi's NMPC solver node's CPU usage percentages for agents 1 and 2 are as follows: peak percentages of 116.2\% and 117.3\%, and average percentages of 98.2\% and 97.82\%, respectively.}
\end{figure}

The resulting trajectories of the HIL simulation are shown in Figure \ref{rpi_traj}, and the time-series data of control inputs are depicted in Figure \ref{input_rpi}. Figure \ref{rpi_solve_time} displays the solving time in milliseconds (ms). The average and peak solving times are 17.19 ms and 95 ms for Agent 1 and 24.6 ms and 95 ms for Agent 2, respectively. The results show that the solver occasionally reached the cut-off time limit, particularly for Agent 2. However, this does not indicate a failure in controlling the robots. Rather, it means that the optimizer did not fully converge to the optimal solution within the allotted time. Nevertheless, the suboptimal solutions returned at the cut-off were sufficiently effective, as demonstrated in Figure \ref{rpi_traj}. Figure \ref{rpi_cpu} shows the CPU usage percentage for the NMPC solver node in each Raspberry Pi. Since it is a quad-core processor, the percentage can be up to 400\%. For Agents 1 and 2, the peak percentages are 116.2\% and 117.3\%, respectively, while the average percentages are 98.2\% and 97.82\%, respectively. This means that the NMPC solver utilized about one core on average.


\section{Conclusion}
In this study, we introduced a local obstacle avoidance strategy tailored to our distributed NMPC-based flock navigation with modified flocking rules. We proposed a point cloud processing technique, including directional filtering and down-sampling, significantly reducing the computational burden. This technique is applicable to both two-dimensional and three-dimensional point cloud data commonly utilized in modern robotic sensors to store obstacle information, such as LiDAR and depth cameras. Then, we introduced a local obstacle avoidance constraint and integrated it into our framework. The real-time solving of the NMPC problem was achieved using PANOC via the OpEn code generator. Simulation results demonstrated the successful navigation of the robot fleet through unknown obstructed environments. Furthermore, an HIL simulation using Raspberry Pi 4 was conducted to assess feasibility. The solving time for the NMPC optimizer was analyzed, revealing peak values reaching the maximum solving time for a few samples, with an average of less than 25 ms. We acknowledge that if the sampling time gets smaller, the chance of hitting the solve time cut-off would increase, which could lead to performance degradation. Thus, further studies can be considered. For example, instead of treating each point in the point cloud separately, we can group nearby points into a single object. This would reduce the number of obstacle constraints from around a hundred to less than ten, which would significantly accelerate the solver and improve real-time performance. On average, the optimizer occupied one core during the HIL simulation. The trajectories in the HIL simulation showed that the robot fleet could navigate through an unknown obstructed environment successfully. However, if the number of detected agents increases, the solving time will also increase, leading to a situation where the optimizer cannot reach the cost value tolerance and undesirable trajectories. Therefore, in addition, future research may explore algorithms that limit the number of agents and prioritize them accordingly.

\section*{Conflict of Interest Statement}

The authors declare that the research was conducted in the absence of any commercial or financial relationships that could be construed as a potential conflict of interest.

\section*{Author Contributions}
NG: Conceptualization, Data curation, Formal analysis, Investigation, Methodology, Software, Validation, Visualization, Writing – original draft, Writing – review and editing. KY: Conceptualization, Formal analysis, Funding acquisition, Project administration, Resources, Supervision, Validation, Writing – original draft, Writing – review and editing.


\section*{Funding}
This work was supported by the Nakajima Foundation and JSPS KAKENHI Grant Number JP24K07546.

\section*{Data Availability Statement}
The UGV model, Husky from \cite{husky}, and the simulated environment by \cite{playen} are used as datasets for validating the proposed algorithm in this study. They can be accessed through the following links: \href{https://clearpathrobotics.com/assets/guides/foxy/husky/index.html}{UGV model} and \href{https://github.com/husky/husky/blob/noetic-devel/husky_gazebo/worlds/clearpath_playpen.world}{Simulated environment}.


\bibliographystyle{Frontiers-Harvard} 
\bibliography{test}

\begin{thebibliography}{29}
\providecommand{\natexlab}[1]{#1}
\expandafter\ifx\csname urlstyle\endcsname\relax
  \providecommand{\doi}[1]{doi:\discretionary{}{}{}#1}\else
  \providecommand{\doi}{doi:\discretionary{}{}{}\begingroup \urlstyle{rm}\Url}\fi
\providecommand{\selectlanguage}[1]{\relax}
\providecommand{\bibAnnoteFile}[1]{%
  \IfFileExists{#1}{\begin{quotation}\noindent\textsc{Key:} #1\\
  \textsc{Annotation:}\ \input{#1}\end{quotation}}{}}
\providecommand{\bibAnnote}[2]{%
  \begin{quotation}\noindent\textsc{Key:} #1\\
  \textsc{Annotation:}\ #2\end{quotation}}

\bibitem[{Andersson et~al.(2019)Andersson, Gillis, Horn, Rawlings, and Diehl}]{casadi}
Andersson, J. A.~E., Gillis, J., Horn, G., Rawlings, J.~B., and Diehl, M. (2019).
\newblock {CasADi: a software framework for nonlinear optimization and optimal control}.
\newblock \emph{Mathematical Programming Computation} 11, 1--36.
\newblock \doi{10.1007/s12532-018-0139-4}
\bibAnnoteFile{casadi}

\bibitem[{Barraquand et~al.(1991)Barraquand, Langlois, and Latombe}]{potential2}
Barraquand, J., Langlois, B., and Latombe, J.-C. (1991).
\newblock Numerical potential field techniques for robot path planning.
\newblock In \emph{Fifth International Conference on Advanced Robotics 'Robots in Unstructured Environments}. 1012--1017 vol.2
\bibAnnoteFile{potential2}

\bibitem[{Brayanov and Stoynova(2019)}]{hil1}
Brayanov, N. and Stoynova, A. (2019).
\newblock Review of hardware-in-the-loop – a hundred years progress in the pseudo-real testing.
\newblock \emph{Electrotechnica \& Electronica} 54, 70--84
\bibAnnoteFile{hil1}

\bibitem[{Cao et~al.(2010)Cao, Chen, Mao, Fang, and Liu}]{reynold_paper1}
Cao, H., Chen, J., Mao, Y., Fang, H., and Liu, H. (2010).
\newblock Formation control based on flocking algorithm in multi-agent system.
\newblock In \emph{2010 8th World Congress on Intelligent Control and Automation}. 2289--2294
\bibAnnoteFile{reynold_paper1}

\bibitem[{{Clearpath Robotics}(2015)}]{husky}
[Dataset] {Clearpath Robotics} (2015).
\newblock Husky {UGV} tutorials 1.0.0.
\newblock Available: https://clearpathrobotics.com
\bibAnnoteFile{husky}

\bibitem[{Goarin et~al.(2024)Goarin, Li, Saviolo, and Loianno}]{dnmpc_cbf}
Goarin, M., Li, G., Saviolo, A., and Loianno, G. (2024).
\newblock Decentralized nonlinear model predictive control for safe collision avoidance in quadrotor teams with limited detection range.
\newblock \emph{arXiv preprint arXiv:2409.17379} IEEE International Conference on Robotics and Automation (ICRA) 2025
\bibAnnoteFile{dnmpc_cbf}

\bibitem[{Koenig and Howard(2004)}]{gazebo}
Koenig, N. and Howard, A. (2004).
\newblock Design and use paradigms for {Gazebo}, an open-source multi-robot simulator.
\newblock In \emph{2004 IEEE/RSJ International Conference on Intelligent Robots and Systems (IROS) (IEEE Cat. No.04CH37566)}. 2149--2154 vol.3
\bibAnnoteFile{gazebo}

\bibitem[{Kong et~al.(2023)Kong, Chen, Li, Yan, Wang, and Fang}]{fixed-wing_flocking}
Kong, F., Chen, H., Li, H., Yan, J., Wang, X., and Fang, J. (2023).
\newblock Flocking with obstacle avoidance for fixed-wing unmanned aerial vehicles via nonlinear model predictive control.
\newblock In \emph{2023 42nd Chinese Control Conference (CCC)}. 5957--5962.
\newblock \doi{10.23919/CCC58697.2023.10240689}
\bibAnnoteFile{fixed-wing_flocking}

\bibitem[{Li et~al.(2024)Li, Yang, Jiang, and Chen}]{reynold_paper2}
Li, C., Yang, Y., Jiang, G., and Chen, X. (2024).
\newblock A flocking control algorithm of multi-agent systems based on cohesion of the potential function.
\newblock \emph{Complex \& Intelligent Systems} 10, 2585--2604
\bibAnnoteFile{reynold_paper2}

\bibitem[{Liang et~al.(2023)Liang, Wang, Yin, Xiong, Zhang, and Yang}]{vfh}
Liang, Q., Wang, Z., Yin, Y., Xiong, W., Zhang, J., and Yang, Z. (2023).
\newblock Autonomous aerial obstacle avoidance using lidar sensor fusion.
\newblock \emph{Plos one} 18, e0287177
\bibAnnoteFile{vfh}

\bibitem[{Lindqvist et~al.(2020)Lindqvist, Mansouri, Kanellakis, and Nikolakopoulos}]{potential3}
Lindqvist, B., Mansouri, S.~S., Kanellakis, C., and Nikolakopoulos, G. (2020).
\newblock Collision free path planning based on local 2d point-clouds for mav navigation.
\newblock In \emph{2020 28th Mediterranean Conference on Control and Automation (MED)}. 538--543
\bibAnnoteFile{potential3}

\bibitem[{Lindqvist et~al.(2021)Lindqvist, Sopasakis, and Nikolakopoulos}]{bjorn1}
Lindqvist, B., Sopasakis, P., and Nikolakopoulos, G. (2021).
\newblock A scalable distributed collision avoidance scheme for multi-agent {UAV} systems.
\newblock In \emph{2021 IEEE/RSJ International Conference on Intelligent Robots and Systems (IROS)}. 9212--9218
\bibAnnoteFile{bjorn1}

\bibitem[{Mestres et~al.(2024)Mestres, Nieto-Granda, and Cortés}]{cbf_1}
Mestres, P., Nieto-Granda, C., and Cortés, J. (2024).
\newblock Distributed safe navigation of multi-agent systems using control barrier function-based controllers.
\newblock \emph{IEEE Robotics and Automation Letters} 9, 6760--6767.
\newblock \doi{10.1109/LRA.2024.3414268}
\bibAnnoteFile{cbf_1}

\bibitem[{Mihalič et~al.(2022)Mihalič, Truntič, and Hren}]{hil2}
Mihalič, F., Truntič, M., and Hren, A. (2022).
\newblock Hardware-in-the-loop simulations: A historical overview of engineering challenges.
\newblock \emph{Electronics} 11, 2462
\bibAnnoteFile{hil2}

\bibitem[{Mukherjee(2015)}]{playen}
[Dataset] Mukherjee, P. (2015).
\newblock clearpath\_playpen.world.
\newblock In husky\_gazebo/worlds, Clearpath Robotics, Inc.
\newblock Available: https://github.com/husky/husky/blob/noetic-devel/husky\_gazebo/worlds/clearpath\_playpen.world
\bibAnnoteFile{playen}

\bibitem[{Nag et~al.(2022)Nag, Huang, Themelis, and Yamamoto}]{aneek1}
Nag, A., Huang, S., Themelis, A., and Yamamoto, K. (2022).
\newblock Flock navigation with dynamic hierarchy and subjective weights using nonlinear {MPC}.
\newblock In \emph{2022 IEEE Conference on Control Technology and Applications (CCTA)}. 1135--1140
\bibAnnoteFile{aneek1}

\bibitem[{Nag and Yamamoto(2024)}]{aneek2}
Nag, A. and Yamamoto, K. (2024).
\newblock Distributed control for flock navigation using nonlinear model predictive control.
\newblock \emph{Advanced Robotics} 38, 619--631
\bibAnnoteFile{aneek2}

\bibitem[{Olfati-Saber(2006)}]{reynold_paper5}
Olfati-Saber, R. (2006).
\newblock Flocking for multi-agent dynamic systems: algorithms and theory.
\newblock \emph{IEEE Transactions on Automatic Control} 51, 401--420.
\newblock \doi{10.1109/TAC.2005.864190}
\bibAnnoteFile{reynold_paper5}

\bibitem[{Reynolds(1987)}]{reynold}
Reynolds, C.~W. (1987).
\newblock Flocks, herds and schools: A distributed behavioral model.
\newblock \emph{SIGGRAPH Comput. Graph.} 21, 25–34.
\newblock \doi{10.1145/37402.37406}
\bibAnnoteFile{reynold}

\bibitem[{Sathya et~al.(2018)Sathya, Sopasakis, Van~Parys, Themelis, Pipeleers, and Patrinos}]{PANOC2}
Sathya, A., Sopasakis, P., Van~Parys, R., Themelis, A., Pipeleers, G., and Patrinos, P. (2018).
\newblock Embedded nonlinear model predictive control for obstacle avoidance using {PANOC}.
\newblock In \emph{2018 European Control Conference (ECC)}. 1523--1528.
\newblock \doi{10.23919/ECC.2018.8550253}
\bibAnnoteFile{PANOC2}

\bibitem[{Shi and Luo(2024)}]{vgf1}
Shi, L. and Luo, J. (2024).
\newblock A framework of point cloud simplification based on voxel grid and its applications.
\newblock \emph{IEEE Sensors Journal} 24, 6349--6357
\bibAnnoteFile{vgf1}

\bibitem[{Song and Kumar(2002)}]{potential1}
Song, P. and Kumar, V. (2002).
\newblock A potential field based approach to multi-robot manipulation.
\newblock In \emph{Proceedings 2002 IEEE International Conference on Robotics and Automation (Cat. No.02CH37292)}. 1217--1222 vol.2
\bibAnnoteFile{potential1}

\bibitem[{Sopasakis et~al.(2020)Sopasakis, Fresk, and Patrinos}]{OpEn}
Sopasakis, P., Fresk, E., and Patrinos, P. (2020).
\newblock {OpEn}: Code generation for embedded nonconvex optimization.
\newblock \emph{IFAC-PapersOnLine} 53, 6548--6554
\bibAnnoteFile{OpEn}

\bibitem[{Stella et~al.(2017)Stella, Themelis, Sopasakis, and Patrinos}]{PANOC}
Stella, L., Themelis, A., Sopasakis, P., and Patrinos, P. (2017).
\newblock A simple and efficient algorithm for nonlinear model predictive control.
\newblock In \emph{IEEE Conference on Decision and Control (CDC)}. 1939--1944
\bibAnnoteFile{PANOC}

\bibitem[{Tanner et~al.(2003{\natexlab{a}})Tanner, Jadbabaie, and Pappas}]{reynold_paper3}
Tanner, H., Jadbabaie, A., and Pappas, G. (2003{\natexlab{a}}).
\newblock Stable flocking of mobile agents, part {I}: fixed topology.
\newblock In \emph{42nd IEEE International Conference on Decision and Control (IEEE Cat. No.03CH37475)}. vol.~2, 2010--2015.
\newblock \doi{10.1109/CDC.2003.1272910}
\bibAnnoteFile{reynold_paper3}

\bibitem[{Tanner et~al.(2003{\natexlab{b}})Tanner, Jadbabaie, and Pappas}]{reynold_paper4}
Tanner, H., Jadbabaie, A., and Pappas, G. (2003{\natexlab{b}}).
\newblock Stable flocking of mobile agents part {II}: dynamic topology.
\newblock In \emph{42nd IEEE International Conference on Decision and Control (IEEE Cat. No.03CH37475)}. vol.~2, 2016--2021.
\newblock \doi{10.1109/CDC.2003.1272911}
\bibAnnoteFile{reynold_paper4}

\bibitem[{Xu et~al.(2023)Xu, Liu, Zhang, Chen, Cui, and Li}]{drone_mpc}
Xu, T., Liu, J., Zhang, Z., Chen, G., Cui, D., and Li, H. (2023).
\newblock Distributed mpc for trajectory tracking and formation control of multi-uavs with leader-follower structure.
\newblock \emph{IEEE Access} 11, 128762--128773.
\newblock \doi{10.1109/ACCESS.2023.3329232}
\bibAnnoteFile{drone_mpc}

\bibitem[{Xu et~al.(2021)Xu, Tong, and Stilla}]{vgf2}
Xu, Y., Tong, X., and Stilla, U. (2021).
\newblock Voxel-based representation of 3d point clouds: Methods, applications, and its potential use in the construction industry.
\newblock \emph{Automation in Construction} 126
\bibAnnoteFile{vgf2}

\bibitem[{Yu et~al.(2021)Yu, Hirche, Huang, Chen, and Allgöwer}]{mpc}
Yu, S., Hirche, M., Huang, Y., Chen, H., and Allgöwer, F. (2021).
\newblock Model predictive control for autonomous ground vehicles: a review.
\newblock \emph{Autonomous Intelligent Systems} 1, 1--17
\bibAnnoteFile{mpc}

\end{thebibliography}


\section*{Figure captions}



\textbf{Figure 1} The illustration of the reference plane in the Directional filtering process.

\noindent \textbf{Figure 2} Agent 2's raw point cloud visualization (left). Agent 2's point cloud visualization after employing Directional filtering (right). A red circle represents the weighted average among three robots, where the position of the leader agent (red triangle) is given more weight.

\noindent \textbf{Figure 3} The Gazebo 3D simulation of three Husky UGVs from \cite{husky} in an obstructed environment by \cite{playen}.

\noindent \textbf{Figure 4} The visualization of laser scan point cloud data from UST10 simulated 2D LiDAR and the robots' frames in RViz.

\noindent \textbf{Figure 5} The schematic diagram of an agent $i$'s distributed NMPC with point cloud processing.

\noindent \textbf{Figure 6} The reference trajectory is shown in red, and the lines connecting waypoints are shown in broken blue. $\text{w}_{i}$ for $i=1, \dotsc 7$ are waypoints.

\noindent \textbf{Figure 7} The trajectories of 3 agents in an obstructed environment, where the blue trajectory is the leader's and the rest are followers.

\noindent \textbf{Figure 8} Time-series control inputs with constraint boundary, shown as broken red lines.

\noindent \textbf{Figure 9} The plot of the deviation from the centroid of the two approaches.

\noindent \textbf{Figure 10} Schematic diagram of the HIL experimentation: The algorithm for the two followers is executed in each Raspberry Pi, while the station computer is responsible for the entire simulation. The information is shared among them using TCP/IP-based client-server communication protocol with ROS.

\noindent \textbf{Figure 11} HIL simulated trajectories of 3 agents in an obstructed environment, where the blue trajectory is the leader's and the rest are followers.

\noindent \textbf{Figure 12} HIL simulation's time-series of the control inputs with boundary, depicted in broken red.

\noindent \textbf{Figure 13} NMPC's solver solving time on Raspberry Pi 4B: Agent 1 averages 17.19 ms and peaks at 95 ms, while Agent 2 averages 24.6 ms and also peaks at 95 ms.

\noindent \textbf{Figure 14} The Raspberry Pi's NMPC solver node's CPU usage percentages for agents 1 and 2 are as follows: peak percentages of 116.2\% and 117.3\%, and average percentages of 98.2\% and 97.82\%, respectively.

\end{document}